\documentclass[letterpaper]{article} 
\usepackage{aaai2027}  
\usepackage[hyphens]{url}  
\usepackage{graphicx} 
\usepackage{natbib}  
\usepackage{caption} 
\usepackage{algorithm}
\usepackage{algorithmic}

\usepackage{newfloat}
\usepackage{listings}

\usepackage{amsmath}
\usepackage{amsfonts}

\usepackage{booktabs}
\usepackage{multirow}

\newcommand{\lr}{\mathbf{x}^{\text{LR}}}
\newcommand{\hr}{\mathbf{x}^{\text{HR}}}
\newcommand{\sr}{\mathbf{x}^{\text{SR}}}

\nocopyright

\DeclareCaptionStyle{ruled}{labelfont=normalfont,labelsep=colon,strut=off} 
\floatstyle{ruled}
\newfloat{listing}{tb}{lst}{}
\floatname{listing}{Listing}
\title{MeanSR: Restoration Trajectory Learning for One-Step Perceptual Super-Resolution}

\author {
    Axi Niu,
    Jiawei Kou,
    Kang Zhang,
    Qingsen Yan,
    Jinqiu Sun,
    Yanning Zhang
}
\affiliations{
    School of Computer Science\\
    Northwestern Polytechnical University\\
    Xi'an, Shaanxi, China\\
    2025202909@mail.nwpu.edu.cn
}

\begin{document}

\maketitle

\begin{abstract}

Diffusion-based super-resolution (SR) achieves strong perceptual quality but requires costly iterative denoising. Existing one-step distillation methods reduce inference time but depend on expensive pretrained teachers, whereas CTMSR avoids distillation through PF-ODE consistency training yet does not explicitly model the restoration dynamics from low-resolution (LR) inputs to high-resolution (HR) images. We propose MeanSR, a one-step perceptual SR method that learns an LR-conditioned average velocity field to directly capture the finite-time transition from degraded or noisy inputs to plausible HR outputs. We further reformulate distribution trajectory matching for average-velocity generation and introduce a Stage-Aware Temporal Sampling strategy to improve trajectory learning. Experiments on synthetic and real-world benchmarks show that MeanSR outperforms CTMSR on CLIPIQA, MUSIQ, and MANIQA while substantially reducing FLOPs and inference latency. MeanSR also reconstructs sharper structures and more realistic textures with fewer perceptual artifacts.
\end{abstract}

\section{Introduction}


Diffusion models have emerged as a leading approach to perceptual single-image super-resolution (SISR) owing to their powerful capability to model complex natural-image distributions and progressively refine noisy observations into perceptually convincing high-resolution (HR) images with realistic high-frequency details~\cite{ho2020ddpm,song2021score,karras2022edm,saharia2022sr3,lin2024diffbir,niu2024acdmsr}.
In particular, StableSR~\cite{wang2024exploiting} and ResShift~\cite{yue2023resshift} have demonstrated impressive restoration quality. However, their iterative denoising processes require multiple network evaluations, resulting in substantial computational cost and inference latency that limit practical deployment.

To alleviate the inefficiency of multi-step sampling, recent studies have explored different one-step diffusion-based super-resolution methods~\cite{wang2024sinsr,wu2024osediff,dong2025tsd,li2025one}. These methods tend to transfer the restoration capability of a multi-step diffusion model to a single-step generator by applying distillation techniques. 
For example, SinSR~\cite{wang2024sinsr} derives a deterministic sampling process from a pretrained diffusion-based SR model and distills the resulting mapping into a one-step network. 
TSD-SR~\cite{dong2025tsd} further improves one-step diffusion SR by introducing target score distillation, which directly transfers the score information of the target HR distribution to guide restoration and enhance perceptual fidelity.
Although these approaches considerably accelerate inference, they still depend on pretrained diffusion priors or teacher models. Such dependence introduces additional training and storage costs, while the performance of the one-step student may be constrained by the teacher model and the quality of its supervision.

\begin{figure}[t]
\centering
\includegraphics[width=0.8\columnwidth]{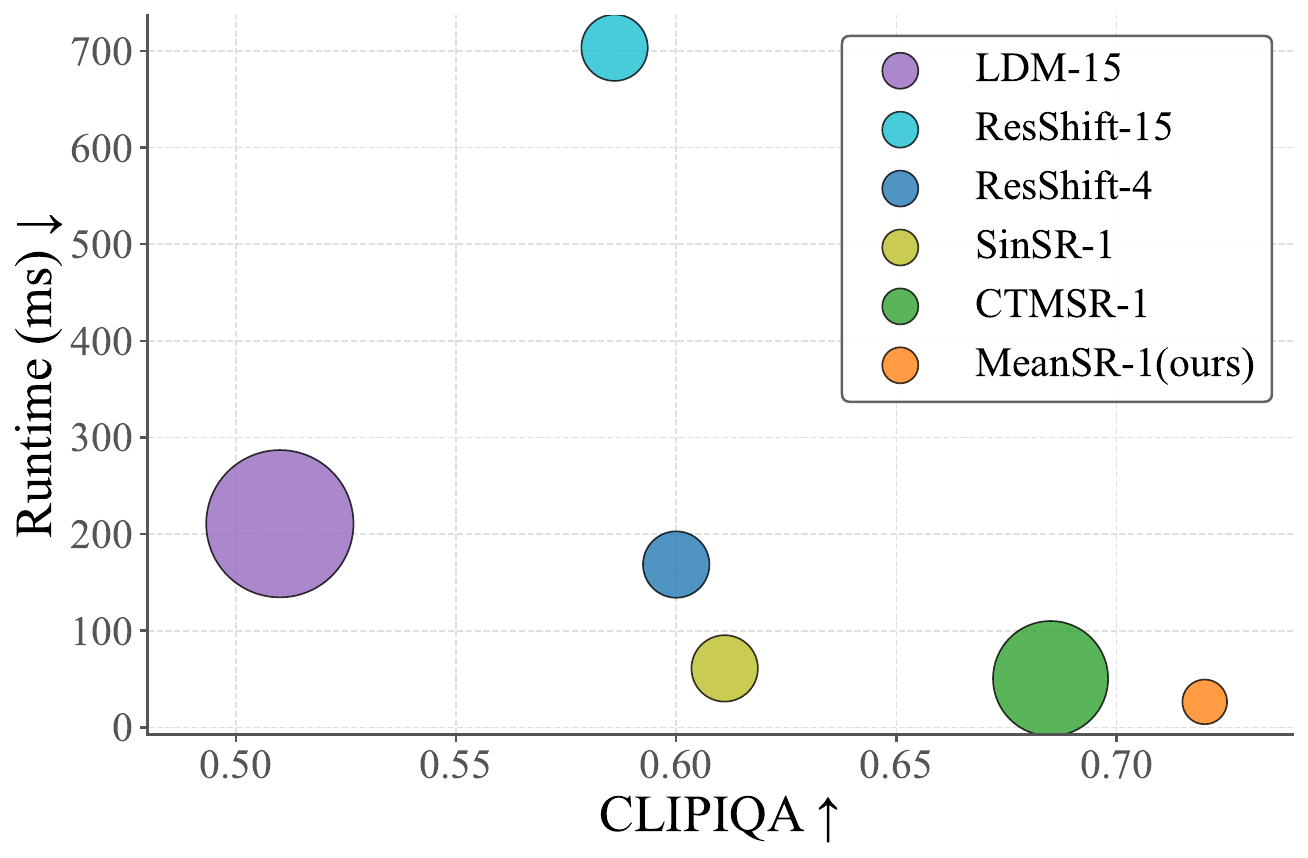}
\caption{Comparison of representative one-step super-resolution methods. The x-axis denotes CLIPIQA, the y-axis denotes inference runtime, and the bubble size represents FLOPs. MeanSR achieves a favorable trade-off between perceptual quality and computational efficiency.}
\label{fig:bubble}
\vspace{-8pt}
\end{figure}

CTMSR~\cite{ctmsr} takes an important step toward addressing these limitations by introducing a distillation-free one-step SR framework. It first constructs a Probability Flow Ordinary Differential Equation (PF-ODE) trajectory from noisy LR observations to HR images and learns the one-step mapping through consistency training~\cite{song2023consistency} by enforcing agreement among predictions at different temporal states. After the consistency training, a Distribution Trajectory Matching (DTM) further align the generated SR distribution with the natural HR distribution to improve the perceptual realism of super resolved images. However, the adopted consistency training process does not explicitly model the underlying restoration field governing the finite-time super resolution process. Consequently, the restoration dynamics remain implicit during optimization, making it difficult to directly supervise trajectory learning and limiting optimization efficiency.


To overcome these limitations, we propose \textbf{MeanSR}, a one-step perceptual SR framework based on LR-conditioned average-velocity learning. Inspired by MeanFlow~\cite{geng2025meanflow}, MeanSR explicitly models the finite-time restoration process by directly estimating the average velocity that transports a degraded and noisy state toward a plausible HR image. The learned average velocity provides an explicit training objective for learning the conditional restoration field and enables one-step restoration through a single network function evaluation (NFE). 
We reformulate the Distribution Trajectory Matching (DTM) objective under the proposed average-velocity generation framework, aligning the generated and target restoration trajectories while preserving explicit restoration field learning. Finally, we introduce \textbf{Stage-Aware Temporal Sampling (SATS)}, which assigns different temporal sampling strategies to restoration trajectory estimation and distribution trajectory matching according to their distinct optimization objectives.

Extensive experiments on synthetic and real-world benchmarks demonstrate that MeanSR achieves superior perceptual quality with substantially reduced computational cost and inference latency. As shown in Fig.~\ref{fig:bubble}, MeanSR achieves the best trade-off between perceptual quality and efficiency among representative one-step super-resolution methods. 


Our main contributions are summarized as follows:

\begin{itemize}
 
\item We propose \textbf{MeanSR}, a one-step perceptual super-resolution framework that explicitly models the finite-time super-resolution field through LR-conditioned average-velocity learning.

\item We reformulate Distribution Trajectory Matching under the average-velocity generation process, providing effective distribution-level supervision for realistic high-frequency detail recovery.

\item We design \textbf{Stage-Aware Temporal Sampling}, which assigns distinct temporal distributions to restoration trajectory estimation and distribution trajectory matching according to their different optimization objectives.

\item A thorough evaluation on synthetic and real-world dataset demonstrate the state-of-the-art performance and faster convergency than existing training based one-step diffusion super-resolution method.

\end{itemize}

\section{Related Work}

\subsection{Diffusion-based Image Super-Resolution}

Recent diffusion models have shown strong generative capability by learning complex data distributions through iterative denoising~\cite{ho2020ddpm,song2021score,karras2022edm}, inspiring their application to single-image super-resolution (SISR). SR3~\cite{saharia2022sr3} pioneers diffusion-based SISR by formulating reconstruction as a conditional denoising process, achieving superior perceptual quality by generating realistic high-frequency details. Subsequent methods, including SRDiff~\cite{li2022srdiff}, ResShift~\cite{yue2023resshift}, ACDMSR~\cite{niu2024acdmsr}, and DiT4SR~\cite{duan2025dit4sr}, further improve restoration quality and sampling efficiency through conditional distribution modeling, residual-based diffusion trajectories, accelerated sampling strategies, and transformer-based architectures. However, these methods still rely on on multiple iterative sampling steps, leading to considerable computational costs and inference latency. This limitation motivates the development of efficient one-step super-resolution frameworks.

\subsection{Diffusion-based One-step Image Super-Resolution}

To overcome the inefficiency of iterative diffusion sampling, recent studies have focused on one-step diffusion-based super-resolution methods~\cite{wang2024sinsr,wu2024osediff,dong2025tsd,li2025one}. SinSR~\cite{wang2024sinsr} distills the sampling trajectory of a pretrained diffusion SR model into a single-step generator, achieving substantially faster inference while largely preserving the restoration capability of the teacher model. However, its performance heavily depends on pretrained diffusion priors and expensive teacher-generated supervision. OSEDiffR~\cite{wu2024osediff} further improves one-step real-world restoration by leveraging latent score distillation, but it likewise relies on large pretrained diffusion models. In contrast, CTMSR~\cite{ctmsr} introduces a distillation-free framework by learning consistency along a Probability Flow Ordinary Differential Equation (PF-ODE) trajectory and further proposes Distribution Trajectory Matching (DTM) to enhance perceptual realism. Although CTMSR successfully removes the dependence on teacher models, it learns restoration through consistency constraints between temporal states, leaving the underlying restoration dynamics implicitly modeled.

\subsection{Trajectory Modeling for One-step Restoration}

Recently, trajectory modeling has emerged as an effective paradigm for efficient image generation. Though  Consistency Models~\cite{song2023consistency} enable one-step or few-step generation by enforcing prediction consistency across different temporal states, they do not explicitly characterize the dynamics governing the generation process. Flow-based methods~\cite{tong2023improving,albergo2025stochastic}, including Flow Matching~\cite{lipman2023flow} and Rectified Flow~\cite{liu2022rectifiedflow}, instead formulate generation as learning transport dynamics between source and target distributions, providing a more explicit representation of the generation trajectory. Building upon this idea, MeanFlow~\cite{geng2025meanflow} models the average velocity over a finite temporal interval, allowing the target sample to be reconstructed through a single prediction without iterative numerical integration. Though these methods have achieved remarkable success in generic image generation, they are not specifically designed for image SR, where structural fidelity and perceptual realism must be jointly preserved. Motivated by these advances, our work extends average-velocity trajectory modeling to one-step perceptual super-resolution.

\begin{figure*}[t]
\centering
\includegraphics[width=\textwidth]{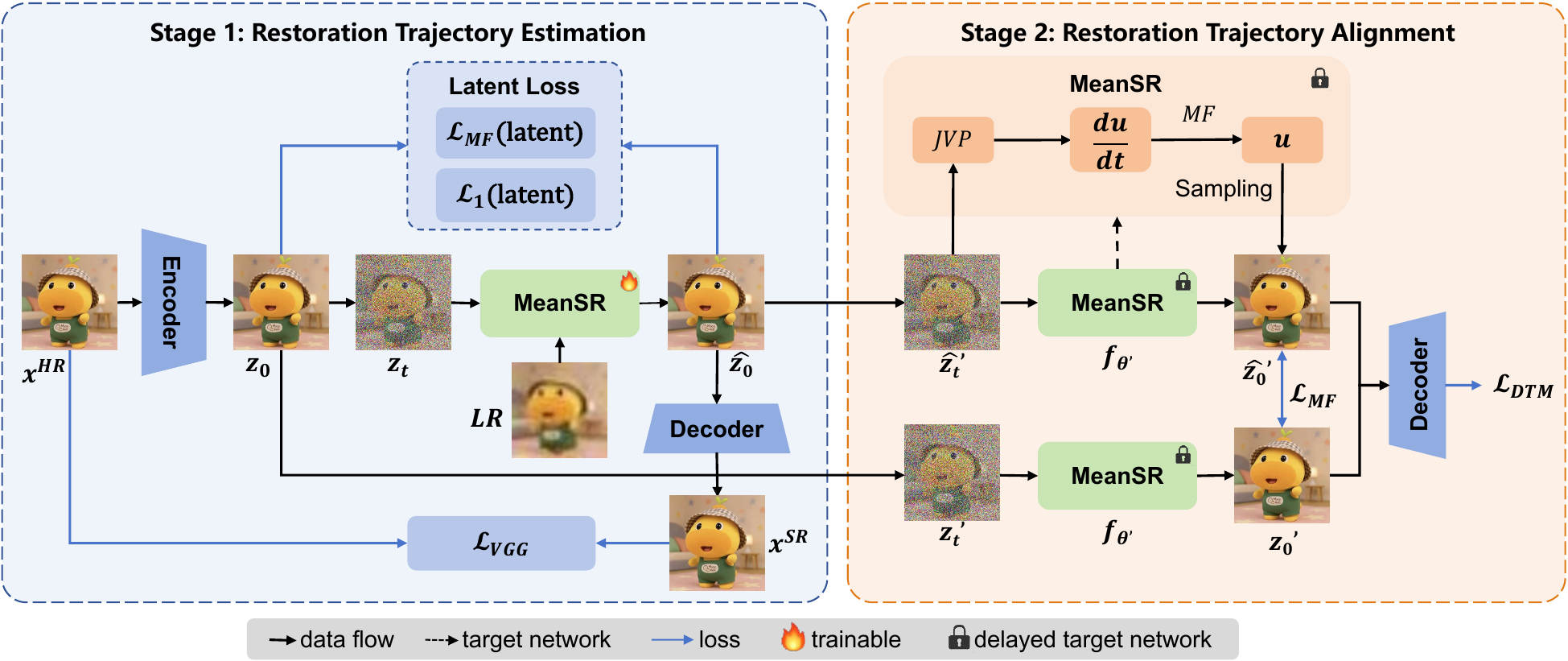}
\caption{Overview of the proposed MeanSR framework. 
Stage 1 estimates an LR-conditioned restoration trajectory by predicting its average velocity under the MeanFlow formulation, while Stage 2 further aligns the generated restoration trajectory with the real HR trajectory through distribution trajectory matching. The proposed Stage-Aware Temporal Sampling (SATS) assigns different temporal distributions to trajectory estimation and trajectory alignment, respectively.
}

\label{fig:meansr_framework}
\end{figure*}
\section{Methodology}

In this section, we present \textbf{MeanSR}, a flow matching based one-step super-resolution framework. As illustrated in Fig.~\ref{fig:meansr_framework}, MeanSR consists of two training stages. In both stages, the LR image is encoded as a conditional feature and concatenated with the noisy latent state as the input, providing LR-guided information for restoration trajectory learning. Stage 1 learns the restoration trajectory by estimating the average velocity that transfers a degraded and noisy state toward a plausible high-resolution (HR) image. Stage 2, termed Distribution Trajectory Matching (DTM), further aligns the generated SR trajectory with the target HR trajectory under the same LR-conditioned restoration process to improve perceptual quality. In addition, a Stage-Aware Temporal Sampling (SATS) strategy is introduced to optimize the two stages with stage-specific temporal distributions.

\subsection{Stage 1: Restoration Trajectory Estimation}

The objective of the first stage is to estimate a one-step HR restoration trajectory under the LR condition. Unlike existing one-step SR methods~\cite{wang2024sinsr,ctmsr} that directly predict the HR endpoint, MeanSR learns the average velocity that guides the flow matching trajectory from one state to another. The learned average velocity directly enables one-step SR generation from random Gaussian noise.

\noindent\textbf{Training.}
Given a clean high resolution image $\hr$ and their degraded low resolution image $\lr$, our target is to generate the SR $\sr$ and train the model to make SR as close to HR as possible $\sr\approx\hr$. We first encode HR and LR into the latent space via the encoder of a pretrained VAE model~\cite{rombach2022high} and obtain $z_0:=z_0^\text{HR}$ and $c^\text{LR}$. As in Flow Matching~\cite{lipman2023flow}, the trajectory point at time step $t$ enables interpolation path:
\begin{equation}
z_t=(1-t)z_0+t\epsilon,
\end{equation}
where $\epsilon\sim\mathcal{N}(0,I)$ and $t\in[0,1]$. Then we aim to learn a conditioned average velocity $u(z_t, c^\text{LR}, r,t)$ to guide the trajectory from samples $z_t$ at time step $t$ directly to target samples $z_r$ at time step $r$:
\begin{equation}
    z_r = z_t - (t-r)u(z_t, c^\text{LR}, r,t)
    \label{eq:sampling}
\end{equation}

To learn the conditioned average velocity $u(z_t, c^\text{LR}, r,t)$, we adopts the MeanFlow formulation~\cite{geng2025meanflow} and defines the average velocity from time step $t$ to $r$ as:
\begin{equation}
u(z_t, c^\text{LR},r,t)
=
\frac{1}{t-r}
\int_r^t
v(z_\tau, c^\text{LR},\tau)\,d\tau,
\label{eq:meanflow_avg_velocity}
\end{equation}
where $v(z_\tau, c^\text{LR},\tau)=\epsilon-z_0$ denotes the instantaneous transport velocity at time step $\tau$.
By differentiating Eq.~\eqref{eq:meanflow_avg_velocity} with respect to $t$ and rearranging the resulting expression, the average velocity $u$ is expressed in terms of the instantaneous velocity $v$ as:
\begin{equation}
u(z_t, c^\text{LR},r,t)
=
v(z_t, c^\text{LR},t)
-
(t-r)\frac{d}{dt}u(z_t, c^\text{LR},r,t),
\label{eq:meanflow_identity}
\end{equation}
where constructing the average velocity requires the temporal derivative of the average velocity $\frac{d}{dt}u$ which is the Jacobian matrix of the function $u$. We use a transformer neural network $u_\theta$ to learn this average velocity $u$ directly:
\begin{equation}
\mathcal{L}_{MeanSR}
=
\mathbb{E}
\|
u_\theta(z_t, c^\text{LR},r,t)
-
\mathrm{sg}(u(z_t, c^\text{LR},r,t))
\|_2^2 ,
\end{equation}

where $\mathrm{sg}(\cdot)$ denotes the stop gradient operation, and according to Equation~\ref{eq:meanflow_identity}, we have $u(z_t, c^\text{LR},r,t)=(\epsilon-z_0) - (t-r)\frac{d}{dt}u_\theta$. Specifically, the Jacobian matrix $\frac{d}{dt}u_\theta$ is efficiently computed using the Jacobian-vector product libraries provided by modern deep learning frameworks such as PyTorch and JAX.

\noindent\textbf{Sampling.}
After training our MeanSR conditional average velocity prediction model $u_\theta(z_t, c^\text{LR},r,t)$, we generate SR $\sr$ from LR $\lr$ in one step via Equation.~\ref{eq:sampling} and simply set the start time step $t=1$ and end time step $r=0$:
\begin{equation}
    z^{\text{SR}} = z_1 - (1-0)u_\theta(z_1, c^\text{LR},0,1),
\end{equation}
where $z_1$ is a randomly samples Gausian noise $z_1\approx\mathcal{N}(0,I)$ and $c^\text{LR}$ is encoded latent space LR. Then we can use the decoder of VAE~\cite{rombach2022high} to easily transfer the $z^{\text{SR}}$ back to pixel space image $\sr$.



\subsection{Stage 2: Restoration Trajectory Alignment}
Stage 2 further aligns the learned super resolution image restoration trajectory with the true data trajectory. We adopt the Distribution Trajectory Matching (DTM) introduced in CTMSR~\cite{ctmsr} and perform the trajectory level distribution alignment rather than endpoint supervision. Specifically, we construct a reference trajectory from the ground-truth latent $z_0$ and compare it with the generated trajectory induced by $\hat{z}_0$ under the same perturbation process.



Given a clean latent $z_0$, we first obtain its noised version at time step $t$, $z_t$, and directly reconstruct the predicted clean latent $\hat{z}_0$ using the average velocity prediction model $u_\theta$. Now, we have a clean latent $z_0$ and a generated latent $\hat{z}_0$.
We then apply the same stochastic perturbation to both $z_0$ and $\hat{z}_0$, yielding paired trajectory states $z_t'$ and $\hat{z}_t'$. These two trajectories are passed through a shared delaying target network to obtain their projected reconstructions $z_0'$ and $\hat{z}_0'$, respectively. The trajectory-level discrepancy is defined as:
\begin{equation}
\Delta = \hat{z}_0' - z_0',
\end{equation}
which captures the misalignment between the generated and ground-truth restoration trajectories.

Based on this discrepancy, we construct a pseudo target that directly corrects the generated trajectory:
\begin{equation}
\tilde{z}_0 = \mathrm{sg}(\hat{z}_0 - \Delta),
\end{equation}
where $\mathrm{sg}(\cdot)$ denotes the stop-gradient operator. This formulation explicitly pushes the generated trajectory toward the true HR trajectory in a self-correcting manner.

To enforce trajectory alignment in the image space, we define the DTM loss as:
\begin{equation}
\mathcal{L}_{DTM} = \mathrm{LPIPS}(D(\hat{z}_0), D(\tilde{z}_0)),
\end{equation}
where $D(\cdot)$ denotes the decoder. This design enforces perceptual consistency at the trajectory level rather than only supervising reconstruction endpoints. During training, the MeanFlow objective is retained to ensure stable trajectory estimation, while DTM provides explicit distribution-level alignment between generated and real restoration trajectories. This leads to improved perceptual realism and more faithful restoration dynamics.

\begin{table*}[!t]
\centering
\resizebox{\textwidth}{!}{
\begin{tabular}{lcccccccccc}
\toprule
\multirow{2}{*}{Method}
& \multicolumn{3}{c}{RealSR}
& \multicolumn{3}{c}{RealSet65}
& \multirow{2}{*}{Steps}
& \multirow{2}{*}{Params (M)}
& \multirow{2}{*}{FLOPs (G)}
& \multirow{2}{*}{Runtime (ms)}
\\

\cmidrule(lr){2-4}
\cmidrule(lr){5-7}

&
CLIPIQA $\uparrow$
& MUSIQ $\uparrow$
& MANIQA $\uparrow$
&
CLIPIQA $\uparrow$
& MUSIQ $\uparrow$
& MANIQA $\uparrow$
&
&
&
&
\\

\midrule

StableSR-200
&0.4124&48.346&0.3021
&0.4488&48.740&0.3097
&200&971.4&44360.844&8244.846
\\

LDM-15
&0.3748&48.698&0.2655
&0.4313&48.602&0.2693
&15&113.6&1269.305&210.767
\\

ResShift-15
&0.5709&\underline{57.769}&0.3691
&0.6309&59.319&0.3916
&15&118.6&101.742&703.378
\\

ResShift-4
&0.5646&55.189&0.3337
&0.6188&58.516&0.3526
&4&118.6&101.742&168.521
\\

SinSR-1
&0.5889&53.145&0.4058
&\underline{0.7164}&62.751&0.4358
&1&118.6&101.742&61.032
\\

CTMSR-1
&\underline{0.6105}&57.283&\underline{0.4574}
&0.6837&\underline{67.183}&\underline{0.4371}
&1&171.5&305.496&50.524
\\

\textbf{MeanSR-1 (ours)}
&\textbf{0.6520}&\textbf{60.171}&\textbf{0.5486}
&\textbf{0.7495}&\textbf{67.498}&\textbf{0.6561}
&1&131.3&\textbf{46.181}&\textbf{26.392}
\\

\bottomrule
\end{tabular}
}
\caption{
Quantitative comparison of perceptual quality and computational efficiency on real-world benchmarks. 
}
\label{tab:real_results_efficiency}

\end{table*}

\begin{table}[!t]
\centering
\resizebox{\columnwidth}{!}{
\begin{tabular}{lccccc}
\toprule
Method & LPIPS $\downarrow$ & CLIPIQA $\uparrow$ & MUSIQ $\uparrow$ & MANIQA $\uparrow$ \\
\midrule
ESRGAN        & 0.485 & 0.451 & 43.615 & 0.3212 \\
BSRGAN        & 0.259 & 0.581 & 54.697 & 0.3865 \\
SwinIR        & 0.238 & 0.564 & 53.790 & 0.3882 \\
\midrule
RealESRGAN    & 0.254 & 0.523 & 52.538 & 0.3689 \\
StableSR-200  & 0.318 & 0.580 & 49.885 & 0.3684 \\
LDM-15        & 0.269 & 0.510 & 46.639 & 0.3305 \\
ResShift-15   & 0.237 & 0.586 & 53.182 & 0.4191 \\
ResShift-4    & \underline{0.208} & 0.600 & 52.019 & 0.3885 \\
\midrule
SinSR-1       & 0.218 & 0.611 & 53.632 & 0.4161 \\
CTMSR-1       & \textbf{0.197} & \underline{0.685} & \underline{59.980} & \underline{0.4846} \\
\textbf{MeanSR-1 (ours)} & 0.228 & \textbf{0.725} & \textbf{61.790} & \textbf{0.5789} \\
\bottomrule
\end{tabular}
}
\caption{Quantitative comparison on the synthetic ImageNet-Test benchmark. }
\label{tab:synthetic_results}
\end{table}

\subsection{Stage-Aware Temporal Sampling (SATS)}
\label{sec:stage_aware_sampling}

To better match the optimization objectives of different training stages, we propose a Stage-Aware Temporal Sampling (SATS) strategy. Instead of adopting a fixed temporal distribution throughout training, SATS assigns stage-specific sampling distributions according to the characteristics of the corresponding learning task. The key motivation of SATS is that the two training stages optimize fundamentally different objectives. The first stage focuses on accurate average-velocity estimation, whereas the second stage aims to align restoration trajectories through DTM. Consequently, different temporal regions contribute unequally to the optimization process, making a unified sampling distribution suboptimal. In the first stage, estimating the average velocity near the data manifold is more critical than learning highly noisy regions. Therefore, SATS adopts a logit-normal distribution, i.e., $t\sim\mathrm{LogNormal}(-0.4,1.0)$, which allocates more samples to low-noise regions and reduces the variance of average velocity estimation, resulting in more stable reconstruction.

In contrast, the second stage focuses on distribution trajectory matching rather than restoration trajectory estimation. Unlike the first stage, which benefits from emphasizing low-noise regions, Distribution Trajectory Matching aligns generated and target restoration trajectories across the entire temporal domain. Consequently, biased temporal supervision leads to incomplete trajectory alignment and weakens distribution-level supervision. Therefore, we adopt uniform temporal sampling, i.e., $t\sim U(0,1)$, which provides balanced supervision over the restoration trajectory and facilitates effective trajectory alignment.

Overall, SATS explicitly couples temporal sampling with the optimization objective of each stage, enabling accurate trajectory estimation in Stage 1 and effective trajectory alignment in Stage 2. This phase-specific design improves perceptual quality without introducing additional inference cost. The complete training algorithm of MeanSR is summarized in Appendix~\ref{App:train_algorithm} in the supplementary material.

\section{Experiments}

\subsection{Experimental Setup}

\paragraph{Training Details.}
We train the proposed MeanSR framework on synthetic super-resolution pairs constructed from the ImageNet dataset. Specifically, high-resolution (HR) images are randomly cropped into $256 \times 256$ patches, and low-resolution (LR) counterparts are generated using the degradation pipeline of RealESRGAN~\cite{wang2021realesrgan}, following common practice in real-world super-resolution. All experiments are conducted under both $\times2$ and $\times4$ super-resolution settings. Please refer to Appendix~\ref{App:train_detail} for more training details. 


\paragraph{Datasets and Evaluation Protocol.}
Although MeanSR is trained solely on synthetic LR-HR pairs, we evaluate its performance on both synthetic and real-world datasets to comprehensively assess its reconstruction accuracy and generalization ability. 
For synthetic evaluation, we construct an ImageNet-Test~\cite{deng2009imagenet} benchmark by randomly selecting 3,000 images from the ImageNet validation set and applying the same degradation process as in training. This setting enables quantitative evaluation with reference-based metrics. 
For real-world evaluation, we adopt two widely used datasets, RealSR~\cite{cai2019toward} and RealSet65~\cite{yue2023resshift}, which contain real captured LR images without exact ground-truth HR counterparts. These datasets are used to assess the generalization capability of the proposed method under practical scenarios.

\paragraph{Evaluation Metrics.}
We evaluate MeanSR using both reference-based and no-reference perceptual quality metrics. For synthetic datasets, we report LPIPS~\cite{zhang2018unreasonable}, which measures perceptual similarity to the ground-truth HR image. For real-world datasets without ground truth, we adopt CLIPIQA~\cite{wang2023exploring}, MUSIQ~\cite{ke2021musiq}, and MANIQA~\cite{yang2022maniqa}, three widely used no-reference metrics that correlate well with human perceptual judgments. These metrics provide a comprehensive evaluation of perceptual realism and visual quality.

\paragraph{Compared Methods.} 
We compare the proposed MeanSR with representative methods from three categories: 
(1) one-step generative methods for super-resolution task, in particular SinSR~\cite{wang2024sinsr} and CTMSR~\cite{ctmsr}; (2) generative super-resolution approaches such as RealESRGAN~\cite{wang2021realesrgan}, 
StableSR~\cite{wang2024exploiting}, LDM~\cite{rombach2022ldm}, and ResShift~\cite{yue2023resshift}; and (3) fidelity-oriented super-resolution methods, including ESRGAN~\cite{wang2018esrgan}, 
BSRGAN~\cite{zhang2021bsrgan}, and SwinIR~\cite{zhang2021swinir}; This diverse set of baselines enables a comprehensive evaluation of perceptual quality and generation realism.

\subsection{Experiment results}

\begin{figure*}[t]
    \centering
    \includegraphics[width=\textwidth]{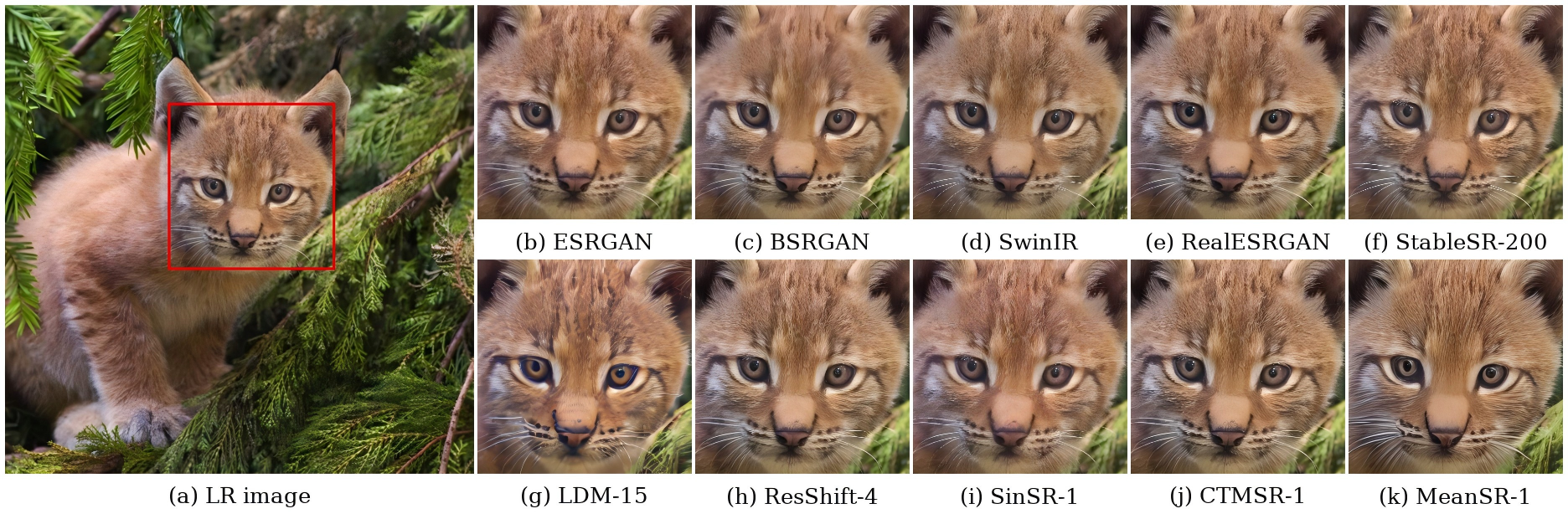}
    \caption{Qualitative comparison on the RealSet65 dataset. Zoom in for a better view. }
    \label{fig:visual_real_cat}
    
\end{figure*}

\begin{figure*}[t]
    \centering
    \includegraphics[width=\textwidth]{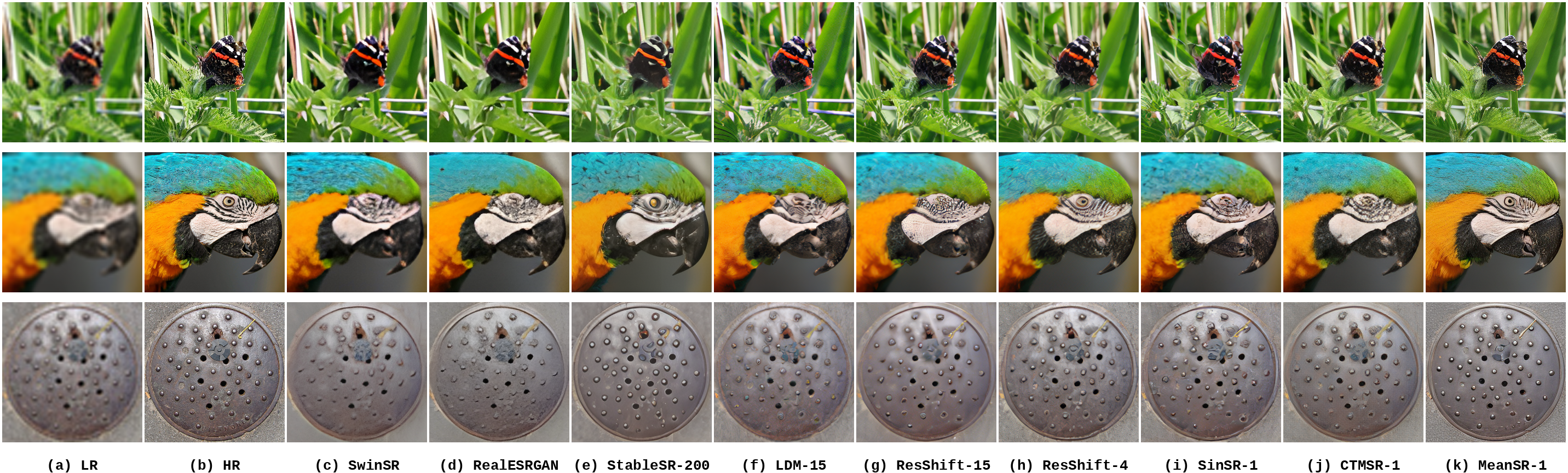}
    \caption{Qualitative comparison on synthetic ImageNet examples. Zoom in for a better view. } 
    \label{fig:visual_synthetic}
\end{figure*}

\paragraph{Quantitative Results on Real-world Data.}
Table~\ref{tab:real_results_efficiency} presents quantitative comparisons on RealSR and RealSet65, including both no-reference perceptual metrics and computational efficiency. MeanSR consistently achieves the best performance across all perceptual metrics on both datasets. Compared with existing multi-step diffusion-based methods, MeanSR obtains substantial improvements in perceptual quality. Specifically, MeanSR improves MANIQA from 0.3021 to 0.5486 on RealSR and from 0.3097 to 0.6561 on RealSet65 over StableSR, corresponding to gains of 81.6\% and 111.9\%, respectively. Compared with LDM, MeanSR further improves MANIQA by 106.8\% on RealSR and 143.6\% on RealSet65, demonstrating the effectiveness of explicit restoration trajectory learning under real-world degradations. Compared with ResShift-15, a representative diffusion-based restoration method, MeanSR achieves 48.6\% and 67.5\% improvements in MANIQA on RealSR and RealSet65, respectively. 

Among one-step methods, MeanSR also consistently outperforms SinSR and CTMSR. Compared with CTMSR, MeanSR improves MANIQA from 0.4574 to 0.5486 on RealSR and from 0.4371 to 0.6561 on RealSet65, corresponding to improvements of 19.9\% and 50.0\%, respectively. These improvements benefit from the proposed LR-conditioned average-velocity learning, which preserves LR-guided structures during restoration trajectory estimation, while Distribution Trajectory Matching further enhances perceptual realism by aligning the generated restoration trajectory with the natural HR distribution. Additional qualitative comparisons are provided in Appendix~\ref{App:Real-world_results}.

In addition to perceptual quality, MeanSR achieves superior computational efficiency. As shown in Table~\ref{tab:real_results_efficiency}, MeanSR requires only 46.18G FLOPs and 26.39 ms inference time, achieving the lowest computational cost among all compared methods. Compared with CTMSR, MeanSR reduces FLOPs by approximately $6\times$ and accelerates inference by nearly $2\times$ while obtaining better perceptual scores. Compared with multi-step diffusion approaches such as StableSR and LDM, MeanSR eliminates iterative denoising and achieves significantly lower latency, demonstrating a favorable balance between perceptual quality and computational efficiency for practical one-step super-resolution.

\paragraph{Quantitative Results on Synthetic Data.}
Table~\ref{tab:synthetic_results} reports quantitative results on ImageNet-Test. Compared with SR methods, generative approaches generally achieve better perceptual quality. Specifically, MeanSR improves MANIQA from 0.3882 of SwinIR to 0.5789, corresponding to a 49.1\% improvement, demonstrating stronger capability in recovering realistic high-frequency details. Compared with multi-step diffusion-based methods, MeanSR also achieves consistent gains over representative approaches such as ResShift-15, improving MANIQA from 0.4191 to 0.5789 (38.1\%) while requiring only one-step inference. Among one-step methods, MeanSR achieves the best CLIPIQA, MUSIQ, and MANIQA scores, surpassing CTMSR by 5.8\%, 3.0\%, and 19.5\%, respectively. Though CTMSR obtains better LPIPS, the superior performance of MeanSR on multiple perceptual metrics indicates stronger perceptual realism and semantic consistency. These results validate the effectiveness of LR-conditioned average-velocity modeling and trajectory-level alignment for one-step perceptual super-resolution.

\begin{figure}[t]
\centering
\includegraphics[width=0.75\columnwidth]{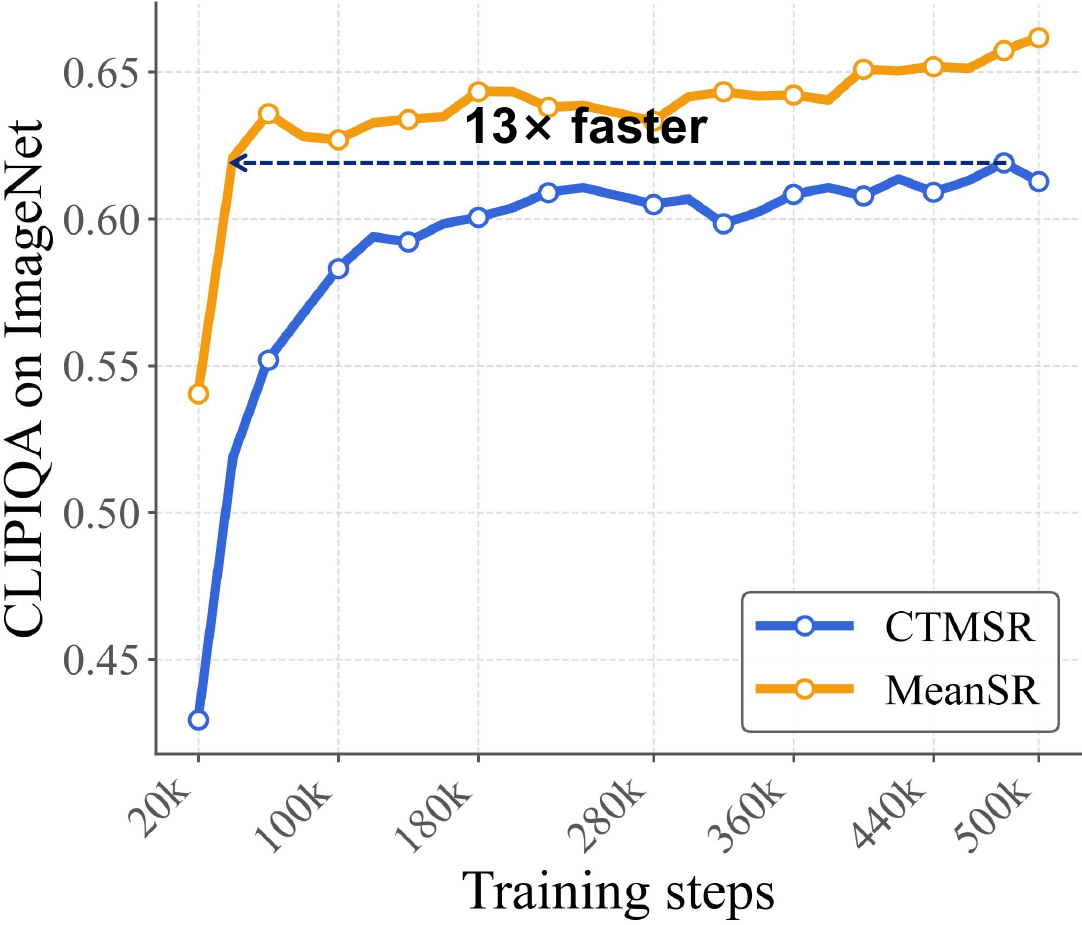}
\caption{Training convergence comparison on ImageNet measured by CLIPIQA. }
\label{fig:convergence}
\end{figure}

We alos analyze the convergence behavior of MeanSR during training on ImageNet. As shown in Fig.~\ref{fig:convergence}, MeanSR reaches a comparable CLIPIQA level with approximately $13\times$ fewer training steps than CTMSR, demonstrating a more efficient convergence behavior. It attributes to the explicit restoration trajectory modeling, which provides direct supervision for learning finite-time restoration dynamics rather than implicitly enforcing consistency among different temporal states. These results further verify the effectiveness of trajectory-based optimization in accelerating the learning process of one-step perceptual super-resolution.


\paragraph{Qualitative Comparison.}
Fig.~\ref{fig:visual_synthetic} and Fig.~\ref{fig:visual_real_cat} provide qualitative comparisons on synthetic and real-world images. MeanSR reconstructs sharper structures and more realistic textures while preserving structural consistency. In the real-world example shown in Fig.~\ref{fig:visual_real_cat}, existing methods often produce over-smoothed fur patterns or introduce unnatural textures in the facial region of the cat. In contrast, MeanSR better recovers fine-grained details, such as individual fur strands, eye contours, and edge structures around the face, resulting in more natural and visually coherent textures. Compared with other one-step methods, MeanSR generates fewer artifacts and maintains better local contrast in detailed regions. These improvements indicate that explicitly modeling the restoration trajectory enables more accurate structure preservation and realistic detail recovery. Overall, the qualitative results are consistent with the quantitative evaluation and further demonstrate the effectiveness of the proposed trajectory learning framework for perceptual super-resolution. Additional qualitative comparisons are provided in Appendix~\ref{App:Synthetic_results}.

\subsection{Ablation Study}

In this section, we conduct experiments to evaluate the effectiveness of Distribution Trajectory Matching (DTM) and different alignment formulations. As shown in Table~\ref{tab:ablation_dtm}, incorporating DTM consistently improves all perceptual metrics. Compared with MeanSR without DTM, applying DTM on the average velocity field improves CLIPIQA from 0.661 to 0.715, MUSIQ from 58.22 to 61.01, and MANIQA from 0.5219 to 0.5761, demonstrating that trajectory-level distribution alignment provides complementary supervision beyond restoration trajectory estimation. Furthermore, performing DTM on the reconstructed image space $\hat{x}_0$ achieves the best performance, improving CLIPIQA to 0.725, MUSIQ to 61.79, and MANIQA to 0.5789.

The superiority of image-space alignment over velocity-space matching indicates that directly optimizing the distribution discrepancy of reconstructed HR images provides more effective perceptual supervision. Although velocity-space DTM constrains the restoration dynamics by aligning the average velocity field, it mainly focuses on the transition process between temporal states and may not fully capture semantic and texture-level differences in the generated images. In contrast, applying DTM on $\hat{x}_0$ directly guides high-frequency detail recovery while preserving the explicit restoration field learning of MeanSR, leading to more realistic and perceptually faithful super-resolution results.

\begin{table}[!t]
\centering
\resizebox{\columnwidth}{!}{

\begin{tabular}{lccc}
\toprule
Methods & CLIPIQA $\uparrow$ & MUSIQ $\uparrow$ & MANIQA $\uparrow$ \\
\midrule
MeanSR (w/o DTM) 
& 0.661 & 58.22 & 0.5219 \\

MeanSR (w/ DTM-$u$) 
& 0.715 & 61.01 & 0.5761 \\

MeanSR (w/ DTM-$x_0$) 
& \textbf{0.725} & \textbf{61.79} & \textbf{0.5789} \\

\bottomrule
\end{tabular}
}

\caption{Distribution Trajectory Matching (DTM) and different alignment formulations on ImageNet-Test.}
\label{tab:ablation_dtm}
\end{table}

\begin{table}[!t]
\centering

\resizebox{\columnwidth}{!}{
\begin{tabular}{lccccc}
\toprule
Steps & LPIPS $\downarrow$ & CLIPIQA $\uparrow$ & MUSIQ $\uparrow$ & MANIQA $\uparrow$ \\
\midrule
10k & \textbf{0.228} & \textbf{0.725} & \textbf{61.79} & \textbf{0.5789} \\
20k & 0.229 & \textbf{0.725} & 60.26 & 0.5502 \\
30k & 0.689 & 0.294 & 23.08 & 0.2822\\
\bottomrule
\end{tabular}
}
\caption{Effect of the number of second-stage training iterations on ImageNet-Test. }
\label{tab:ablation_steps}
\end{table}

\begin{table}[!t]
\centering

\resizebox{\columnwidth}{!}{
\begin{tabular}{ccccc}
\toprule
Stage & Sampling & CLIPIQA $\uparrow$ & MUSIQ $\uparrow$ & MANIQA $\uparrow$ \\
\midrule

\multirow{2}{*}{Stage 1}
& $U(0,1)$ & 0.6317 & 56.2622 & 0.4861 \\
& $\mathrm{logN}(-0.4,1.0)$ & \textbf{0.6618} & \textbf{58.2219} & \textbf{0.5219} \\

\midrule

\multirow{6}{*}{Stage 2}
& $U(0.1,0.5)$ & 0.7110 & 61.35 & 0.5384 \\
& $U(0,1)$ & \textbf{0.7250} & \textbf{61.79} & \textbf{0.5789} \\
& $U(0.5,1)$ & 0.7240 & 61.56 & 0.5774 \\
& $U(0,0.1)$ & 0.6948 & 59.78 & 0.4931 \\
& $U(0.9,1)$ & 0.6411 & 53.42 & 0.4391 \\
& $\mathrm{logN}(-0.4,1.0)$ & 0.6907 & 59.39 & 0.4856 \\

\bottomrule
\end{tabular}
}
\caption{Ablation of the proposed Stage-Aware Temporal Sampling (SATS). Stage 2 is initialized from the Stage 1 model pretrained with logit-normal sampling.}
\label{tab:ablation_time}
\end{table}



Then, we investigate the effect of the second-stage optimization length. Table~\ref{tab:ablation_steps} reveals the impact of the second-stage optimization length. The best performance is achieved after 10k fine-tuning steps, while longer optimization gradually degrades perceptual quality, indicating that excessive trajectory alignment over-constrain the generation process.

We further evaluate the effectiveness of the proposed SATS strategy in Table~\ref{tab:ablation_time}. In the Restoration Trajectory Estimation stage, logit-normal sampling achieves better performance than uniform sampling, demonstrating that emphasizing low-noise regions provides more stable supervision for average-velocity estimation. Based on the optimized Stage~1 model with logit-normal sampling, we further compare different temporal sampling strategies for the Distribution Trajectory Matching stage. The results show that uniform sampling over the entire temporal range $U(0,1)$ achieves the best performance, while biased temporal sampling weakens trajectory alignment by providing incomplete supervision across the restoration process. These results verify that the two optimization stages have distinct temporal requirements and validate the effectiveness of the proposed stage-aware temporal sampling strategy.


\section{Conclusion}

In this paper, we proposed MeanSR, a degradation-aware one-step generative super-resolution framework that formulates perceptual super-resolution as a restoration trajectory learning problem. By combining LR-conditioned average-velocity modeling, Distribution Trajectory Matching (DTM), and Stage-Aware Temporal Sampling (SATS), MeanSR effectively models restoration dynamics while maintaining distribution-level perceptual supervision. Extensive experiments on synthetic and real-world benchmarks demonstrate that MeanSR consistently achieves superior perceptual quality with substantially lower computational cost and inference latency than existing one-step methods. These results highlight restoration trajectory learning as an effective paradigm for one-step perceptual super-resolution. Future work will explore more expressive degradation-aware trajectory modeling and its extension to other image restoration tasks.

\clearpage

\bibliography{aaai2027}

@inproceedings{ho2020ddpm,
  title={Denoising Diffusion Probabilistic Models},
  author={Ho, Jonathan and Jain, Ajay and Abbeel, Pieter},
  booktitle={Advances in Neural Information Processing Systems},
  volume={33},
  pages={6840--6851},
  year={2020}
}

@inproceedings{song2021score,
  title={Score-Based Generative Modeling through Stochastic Differential Equations},
  author={Song, Yang and Sohl-Dickstein, Jascha and Kingma, Diederik P. and Kumar, Abhishek and Ermon, Stefano and Poole, Ben},
  booktitle={International Conference on Learning Representations},
  year={2021}
}

@inproceedings{karras2022edm,
  title={Elucidating the Design Space of Diffusion-Based Generative Models},
  author={Karras, Tero and Aittala, Miika and Aila, Timo and Laine, Samuli},
  booktitle={Advances in Neural Information Processing Systems},
  year={2022}
}

@article{song2023consistency,
  title={Consistency models},
  author={Song, Yang and Dhariwal, Prafulla and Chen, Mark and Sutskever, Ilya}
}

@inproceedings{lipman2023flow,
  title={Flow Matching for Generative Modeling},
  author={Lipman, Yaron and Chen, Ricky T. Q. and Ben-Hamu, Heli and Nickel, Maximilian and Le, Matt},
  booktitle={International Conference on Learning Representations},
  year={2023}
}

@inproceedings{liu2022rectifiedflow,
  title={Flow Straight and Fast: Learning to Generate and Transfer Data with Rectified Flow},
  author={Liu, Xingchao and Gong, Chengyue and Liu, Qiang},
  booktitle={International Conference on Learning Representations},
  year={2023}
}

@article{geng2025meanflow,
  title={Mean Flows for One-Step Generative Modeling},
  author={Geng, Zhengyang and Deng, Mingyang and Bai, Xingjian and Kolter, J. Zico and He, Kaiming},
  journal={arXiv preprint arXiv:2505.13447},
  year={2025}
}

@inproceedings{ctmsr,
  title={Consistency Trajectory Matching for One-Step Generative Super-Resolution},
  author={You, Weiyi and Zhang, Mingyang and Zhang, Leheng and Zhou, Xingyu and Shi, Kexuan and Gu, Shuhang},
  booktitle={Proceedings of the IEEE/CVF International Conference on Computer Vision},
  pages={12747--12756},
  year={2025}
}

@inproceedings{wang2018esrgan,
  title={{ESRGAN}: Enhanced Super-Resolution Generative Adversarial Networks},
  author={Wang, Xintao and Yu, Ke and Wu, Shixiang and Gu, Jinjin and Liu, Yihao and Dong, Chao and Qiao, Yu and Loy, Chen Change},
  booktitle={Proceedings of the European Conference on Computer Vision Workshops},
  year={2018}
}

@inproceedings{zhang2021swinir,
  title={{SwinIR}: Image Restoration Using Swin Transformer},
  author={Zhang, Jingyun and Zeng, Huan and Guo, Yuchao and Zhang, Lei},
  booktitle={Proceedings of the IEEE/CVF International Conference on Computer Vision Workshops},
  year={2021}
}

@inproceedings{wang2021realesrgan,
  title={Real-{ESRGAN}: Training Real-World Blind Super-Resolution with Pure Synthetic Data},
  author={Wang, Xintao and Xie, Liangbin and Dong, Chao and Shan, Ying},
  booktitle={Proceedings of the IEEE/CVF International Conference on Computer Vision Workshops},
  year={2021}
}

@inproceedings{rombach2022ldm,
  title={High-Resolution Image Synthesis with Latent Diffusion Models},
  author={Rombach, Robin and Blattmann, Andreas and Lorenz, Dominik and Esser, Patrick and Ommer, Bjorn},
  booktitle={Proceedings of the IEEE/CVF Conference on Computer Vision and Pattern Recognition},
  pages={10684--10695},
  year={2022}
}

@article{wang2024exploiting,
  title={Exploiting diffusion prior for real-world image super-resolution},
  author={Wang, Jianyi and Yue, Zongsheng and Zhou, Shangchen and Chan, Kelvin CK and Loy, Chen Change},
  journal={International Journal of Computer Vision},
  volume={132},
  number={12},
  pages={5929--5949},
  year={2024},
  publisher={Springer}
}

@inproceedings{yue2023resshift,
  title={ResShift: Efficient Diffusion Model for Image Super-Resolution by Residual Shifting},
  author={Yue, Zongsheng and Wang, Jianyi and Loy, Chen Change},
  booktitle={NeurIPS},
  year={2023}
}

@inproceedings{zhang2021bsrgan,
    title={Designing a Practical Degradation Model for Deep Blind Image Super-Resolution},
    author={Zhang, Kai and Liang, Jingyun and Van Gool, Luc and Timofte, Radu},
    booktitle={Proceedings of the IEEE/CVF international conference on computer vision},
    pages={4791--4800},
    year={2021}
}

@inproceedings{wang2024sinsr,
  title={SinSR: diffusion-based image super-resolution in a single step},
  author={Wang, Yufei and Yang, Wenhan and Chen, Xinyuan and Wang, Yaohui and Guo, Lanqing and Chau, Lap-Pui and Liu, Ziwei and Qiao, Yu and Kot, Alex C and Wen, Bihan},
  booktitle={Proceedings of the IEEE/CVF Conference on Computer Vision and Pattern Recognition},
  pages={25796--25805},
  year={2024}
}

@article{niu2024acdmsr,
  title={ACDMSR: Accelerated conditional diffusion models for single image super-resolution},
  author={Niu, Axi and Pham, Trung X and Zhang, Kang and Sun, Jinqiu and Zhu, Yu and Yan, Qingsen and Kweon, In So and Zhang, Yanning},
  journal={IEEE Transactions on Broadcasting},
  volume={70},
  number={2},
  pages={492--504},
  year={2024},
  publisher={IEEE}
}

@article{saharia2022sr3,
  title={Image super-resolution via iterative refinement},
  author={Saharia, Chitwan and Ho, Jonathan and Chan, William and Salimans, Tim and Fleet, David J and Norouzi, Mohammad},
  journal={IEEE transactions on pattern analysis and machine intelligence},
  volume={45},
  number={4},
  pages={4713--4726},
  year={2022},
  publisher={IEEE}
}

@inproceedings{lin2024diffbir,
  title={Diffbir: Toward blind image restoration with generative diffusion prior},
  author={Lin, Xinqi and He, Jingwen and Chen, Ziyan and Lyu, Zhaoyang and Dai, Bo and Yu, Fanghua and Qiao, Yu and Ouyang, Wanli and Dong, Chao},
  booktitle={European conference on computer vision},
  pages={430--448},
  year={2024},
  organization={Springer}
}

@inproceedings{wu2024osediff,
  title={One-Step Effective Diffusion Network for Real-World Image Super-Resolution},
  author={Wu, Rongyuan and Sun, Lingchen and Ma, Zhiyuan and Zhang, Lei},
  booktitle={Advances in Neural Information Processing Systems},
  year={2024}
}

@inproceedings{dong2025tsd,
  title={Tsd-sr: One-step diffusion with target score distillation for real-world image super-resolution},
  author={Dong, Linwei and Fan, Qingnan and Guo, Yihong and Wang, Zhonghao and Zhang, Qi and Chen, Jinwei and Luo, Yawei and Zou, Changqing},
  booktitle={Proceedings of the Computer Vision and Pattern Recognition Conference},
  pages={23174--23184},
  year={2025}
}

@article{li2025one,
  title={One diffusion step to real-world super-resolution via flow trajectory distillation},
  author={Li, Jianze and Cao, Jiezhang and Guo, Yong and Li, Wenbo and Zhang, Yulun},
  journal={arXiv preprint arXiv:2502.01993},
  year={2025}
}

@article{li2022srdiff,
  title={Srdiff: Single image super-resolution with diffusion probabilistic models},
  author={Li, Haoying and Yang, Yifan and Chang, Meng and Chen, Shiqi and Feng, Huajun and Xu, Zhihai and Li, Qi and Chen, Yueting},
  journal={Neurocomputing},
  volume={479},
  pages={47--59},
  year={2022},
  publisher={Elsevier}
}

@article{tong2023improving,
  title={Improving and generalizing flow-based generative models with minibatch optimal transport},
  author={Tong, Alexander and Fatras, Kilian and Malkin, Nikolay and Huguet, Guillaume and Zhang, Yanlei and Rector-Brooks, Jarrid and Wolf, Guy and Bengio, Yoshua},
  journal={arXiv preprint arXiv:2302.00482},
  year={2023}
}

@article{albergo2025stochastic,
  title={Stochastic interpolants: A unifying framework for flows and diffusions},
  author={Albergo, Michael and Boffi, Nicholas M and Vanden-Eijnden, Eric},
  journal={Journal of Machine Learning Research},
  volume={26},
  number={209},
  pages={1--80},
  year={2025}
}

@inproceedings{duan2025dit4sr,
  title={Dit4sr: Taming diffusion transformer for real-world image super-resolution},
  author={Duan, Zheng-Peng and Zhang, Jiawei and Jin, Xin and Zhang, Ziheng and Xiong, Zheng and Zou, Dongqing and Ren, Jimmy S and Guo, Chunle and Li, Chongyi},
  booktitle={Proceedings of the IEEE/CVF International Conference on Computer Vision},
  pages={18948--18958},
  year={2025}
}

@inproceedings{deng2009imagenet,
  title={Imagenet: A large-scale hierarchical image database},
  author={Deng, Jia and Dong, Wei and Socher, Richard and Li, Li-Jia and Li, Kai and Fei-Fei, Li},
  booktitle={2009 IEEE conference on computer vision and pattern recognition},
  pages={248--255},
  year={2009},
  organization={Ieee}
}

@inproceedings{cai2019toward,
  title={Toward real-world single image super-resolution: A new benchmark and a new model},
  author={Cai, Jianrui and Zeng, Hui and Yong, Hongwei and Cao, Zisheng and Zhang, Lei},
  booktitle={Proceedings of the IEEE/CVF international conference on computer vision},
  pages={3086--3095},
  year={2019}
}

@inproceedings{rombach2022high,
  title={High-resolution image synthesis with latent diffusion models},
  author={Rombach, Robin and Blattmann, Andreas and Lorenz, Dominik and Esser, Patrick and Ommer, Bj{\"o}rn},
  booktitle={Proceedings of the IEEE/CVF conference on computer vision and pattern recognition},
  pages={10684--10695},
  year={2022}
}

@inproceedings{wang2023exploring,
  title={Exploring clip for assessing the look and feel of images},
  author={Wang, Jianyi and Chan, Kelvin CK and Loy, Chen Change},
  booktitle={Proceedings of the AAAI conference on artificial intelligence},
  volume={37},
  number={2},
  pages={2555--2563},
  year={2023}
}

@inproceedings{ke2021musiq,
  title={Musiq: Multi-scale image quality transformer},
  author={Ke, Junjie and Wang, Qifei and Wang, Yilin and Milanfar, Peyman and Yang, Feng},
  booktitle={Proceedings of the IEEE/CVF international conference on computer vision},
  pages={5148--5157},
  year={2021}
}

@inproceedings{yang2022maniqa,
  title={Maniqa: Multi-dimension attention network for no-reference image quality assessment},
  author={Yang, Sidi and Wu, Tianhe and Shi, Shuwei and Lao, Shanshan and Gong, Yuan and Cao, Mingdeng and Wang, Jiahao and Yang, Yujiu},
  booktitle={Proceedings of the IEEE/CVF conference on computer vision and pattern recognition},
  pages={1191--1200},
  year={2022}
}

@inproceedings{zhang2018unreasonable,
  title={The unreasonable effectiveness of deep features as a perceptual metric},
  author={Zhang, Richard and Isola, Phillip and Efros, Alexei A and Shechtman, Eli and Wang, Oliver},
  booktitle={Proceedings of the IEEE conference on computer vision and pattern recognition},
  pages={586--595},
  year={2018}
}

\clearpage
\newpage

\newcommand{\code}[1]{\textrm{#1}}
\newcommand{\fn}[1]{\textrm{#1}}
\newcommand{\kw}[1]{\textbf{#1}}
\newcommand{\cmt}[1]{\textrm{\emph{#1}}}

\appendix








\section{Training details}
\label{App:train_detail}

The model is trained on 4 NVIDIA A100 GPUs with a total batch size of 32. The entire training process takes approximately 100 hours. Following the two-stage framework of MeanSR, training is divided into two stages. In the first stage, the model is trained for 500k iterations using the MeanFlow objective together with pixel-wise $\ell_1$ loss and perceptual VGG loss to learn a stable restoration trajectory. In the second stage, the model is fine-tuned for an additional 10k iterations with the DTM objective, which aligns the generated trajectory with the target HR trajectory.

The loss weights are empirically set to 0.5 for the $\ell_1$ loss, 0.1 for the VGG loss, and 1.6 for the DTM loss. We find that a relatively short DTM fine-tuning stage is sufficient to substantially improve perceptual quality while preserving the stability of the learned restoration trajectory.

All quantitative results are reported from a single training run under fixed experimental settings.

\section{Training Algorithm}
\label{App:train_algorithm}

Based on the framework described in the main paper, the complete training procedure of MeanSR consists of two sequential optimization stages with stage-aware temporal sampling. In both stages, the LR image is first encoded into a conditional feature and concatenated with the noisy latent state as the input of MeanSR. The network then predicts the LR-conditioned average velocity field, which characterizes the finite-time transition from the degraded state to the target HR image.

In the first stage, namely Restoration Trajectory Estimation, the objective is to learn a stable restoration trajectory through the MeanFlow formulation. Given an HR image $x_0$ and its corresponding LR condition $y$, a noisy state $x_t$ is constructed by interpolating between the HR image and Gaussian noise. To obtain more stable average velocity estimation, the timestep is sampled using a logit-normal distribution, which assigns more training ssamples to relatively low-noise regions. The MeanSR network predicts the average velocity field $u_\theta(x_t,y,r,t)$, and the MeanFlow objective is optimized by matching the predicted velocity with the target velocity derived from the flow trajectory. In addition, pixel-wise $\ell_1$ loss and perceptual VGG loss are applied to the reconstructed HR estimation to preserve structural fidelity and perceptual details. The complete optimization procedure of Stage~1 is summarized in Algorithm~\ref{alg:stage1}.

After obtaining a stable restoration trajectory, the second stage further improves the perceptual quality through Distribution Trajectory Matching (DTM). Different from the first stage, which focuses on accurate velocity estimation, the second stage aims to align the generated restoration trajectory with the target HR trajectory. Therefore, uniform temporal sampling is adopted to provide balanced supervision across the entire temporal domain. During this stage, the predicted SR trajectory is compared with the target HR trajectory through the DTM objective, while the MeanFlow loss is retained to maintain the learned restoration dynamics. The detailed optimization procedure of Stage~2 is provided in Algorithm~\ref{alg:stage2}.

Through the two-stage optimization strategy, MeanSR first learns an explicit LR-conditioned restoration field and then refines the generated trajectory distribution. The stage-aware temporal sampling strategy enables each stage to focus on its corresponding optimization objective, resulting in more stable trajectory learning and improved perceptual restoration quality.

\begin{algorithm}[t]
\caption{Stage 1: Restoration Trajectory Estimation}
\label{alg:stage1}

\cmt{\# fn(x\_t, y, r, t): predict MeanSR velocity field u\_theta}\\
\cmt{\# x\_0: HR image, y: LR condition}\\[1mm]

\code{t, r = }\fn{sample\_lognormal}\code{(-0.4, 1.0)}\\
\code{e = }\fn{randn\_like}\code{(x\_0)}\\[1mm]

\code{x\_t = (1 - t) * x\_0 + t * e}\\
\code{v = e - x\_0}\\[1mm]

\code{u, dudt = }\fn{jvp}\code{(fn, (x\_t, y, r, t), (v, 0, 0, 1))}\\[1mm]

\code{u\_tgt = v - (t - r) * dudt}\\
\code{x\_hat\_0 = x\_t - (t - r) * u}\\[1mm]

\code{loss\_mf = }\fn{metric}\code{(u - }\fn{stopgrad}\code{(u\_tgt))}\\
\code{loss\_l1 = }\fn{l1\_loss}\code{(x\_hat\_0, x\_0)}\\
\code{loss\_vgg = }\fn{vgg\_loss}\code{(x\_hat\_0, x\_0)}\\[1mm]

\code{loss = loss\_mf + lambda1 * loss\_l1 + lambda2 * loss\_vgg}\\[1mm]

\vspace{1mm}

\end{algorithm}

\begin{algorithm}[t]
\caption{Stage 2: Restoration Trajectory Alignment}
\label{alg:stage2}

\cmt{\# target\_step(): delayed target-network projection}\\[1mm]

\kw{if} \code{step \% 1000 == 0:}\\
\hspace*{4mm}\fn{target.load\_state\_dict}\code{(online.}\fn{state\_dict}\code{())}\\[1mm]

\code{t, r = }\fn{sample\_uniform}\code{(0,1)}\\
\code{e = }\fn{randn\_like}\code{(x\_0)}\\[1mm]

\code{x\_t = (1 - t) * x\_0 + t * e}\\
\code{v = e - x\_0}\\[1mm]

\code{u, dudt = }\fn{jvp}\code{(fn, (x\_t, y, r, t), (v, 0, 0, 1))}\\
\code{u\_tgt = v - (t - r) * dudt}\\
\code{x\_hat\_0 = x\_t - (t - r) * u}\\[1mm]

\code{x\_hat\_t, x\_t\_real = }\fn{perturb}\code{(x\_hat\_0, x\_0, t)}\\
\code{x\_hat\_0\_t, x\_0\_t = }\fn{target\_step}\code{(x\_hat\_t, x\_t\_real, y, r, t)}\\
\code{grad = x\_hat\_0\_t - x\_0\_t}\\[1mm]

\code{loss\_dtm = 0.5 * }\fn{lpips}\code{(x\_hat\_0, }\fn{stopgrad}\code{(x\_hat\_0 - grad))}\\
\code{loss\_mf = }\fn{metric}\code{(u - }\fn{stopgrad}\code{(u\_tgt))}\\
\code{loss = loss\_mf + lambda\_dtm * loss\_dtm}\\[1mm]

\vspace{1mm}
\end{algorithm}


\section{Visual Results on Real-world Datasets}
\label{App:Real-world_results}

\begin{figure*}[t]
    \centering
    \includegraphics[width=\textwidth]{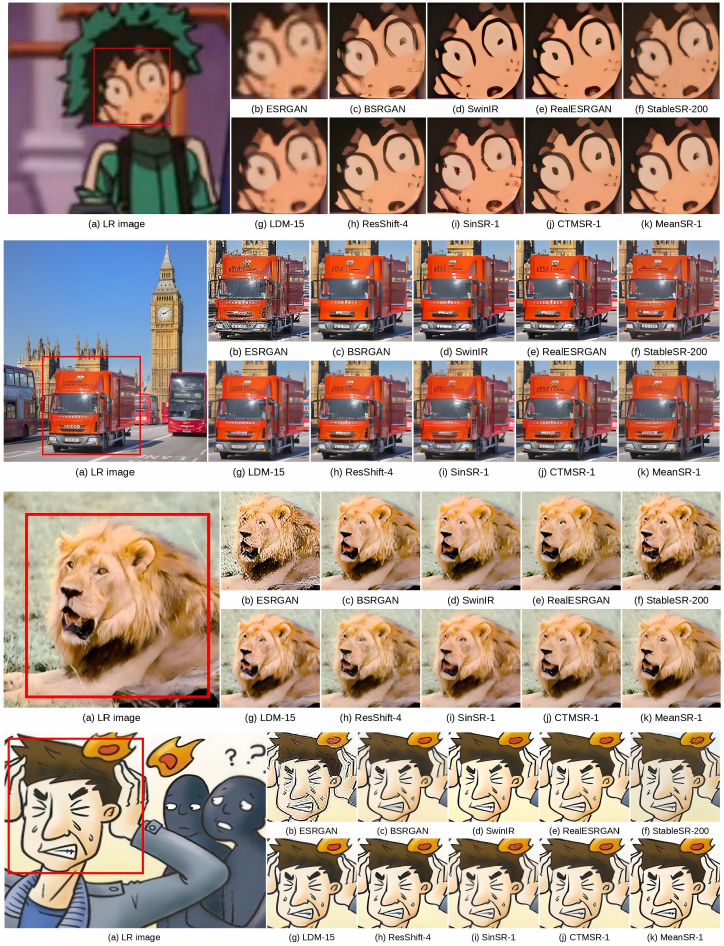}
    \caption{Qualitative comparison on the RealSet65 dataset. }
    \label{fig:supp_realset65}
\end{figure*}

\begin{figure*}[t]
    \centering
    \includegraphics[width=\textwidth]{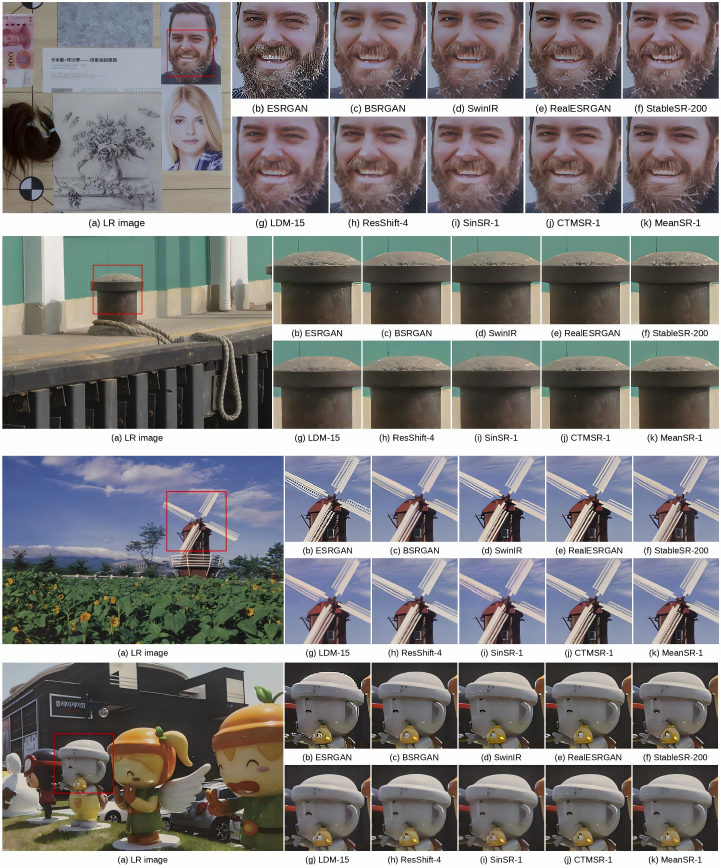}
    \caption{Qualitative comparison on the RealSR dataset. }
    \label{fig:supp_realsr}
\end{figure*}

We provide additional qualitative comparisons on real-world super-resolution benchmarks, including RealSet65 and RealSR. As shown in Fig.~\ref{fig:supp_realset65} and Fig.~\ref{fig:supp_realsr}, MeanSR consistently generates visually more realistic high-resolution images compared with both reconstruction-based and generative super-resolution methods.

Compared with conventional reconstruction-oriented approaches, MeanSR recovers sharper edges, clearer object structures, and more natural textures, while effectively avoiding over-smoothed artifacts caused by pixel-wise optimization. In challenging regions containing fine details, repetitive patterns, and complex textures, MeanSR preserves more accurate local structures and produces visually plausible high-frequency details.

Compared with diffusion-based multi-step methods, MeanSR achieves comparable or superior perceptual quality with only one-step inference by explicitly learning the LR-conditioned restoration trajectory. Moreover, compared with existing one-step approaches, MeanSR generates fewer artifacts and better maintains semantic consistency. These visual results further demonstrate that the proposed average-velocity modeling and Distribution Trajectory Matching effectively guide perceptual restoration under real-world degradations.

\section{Visual Results on Synthetic Dataset}
\label{App:Synthetic_results}

\begin{figure*}[t]
    \centering
    \includegraphics[width=\textwidth]{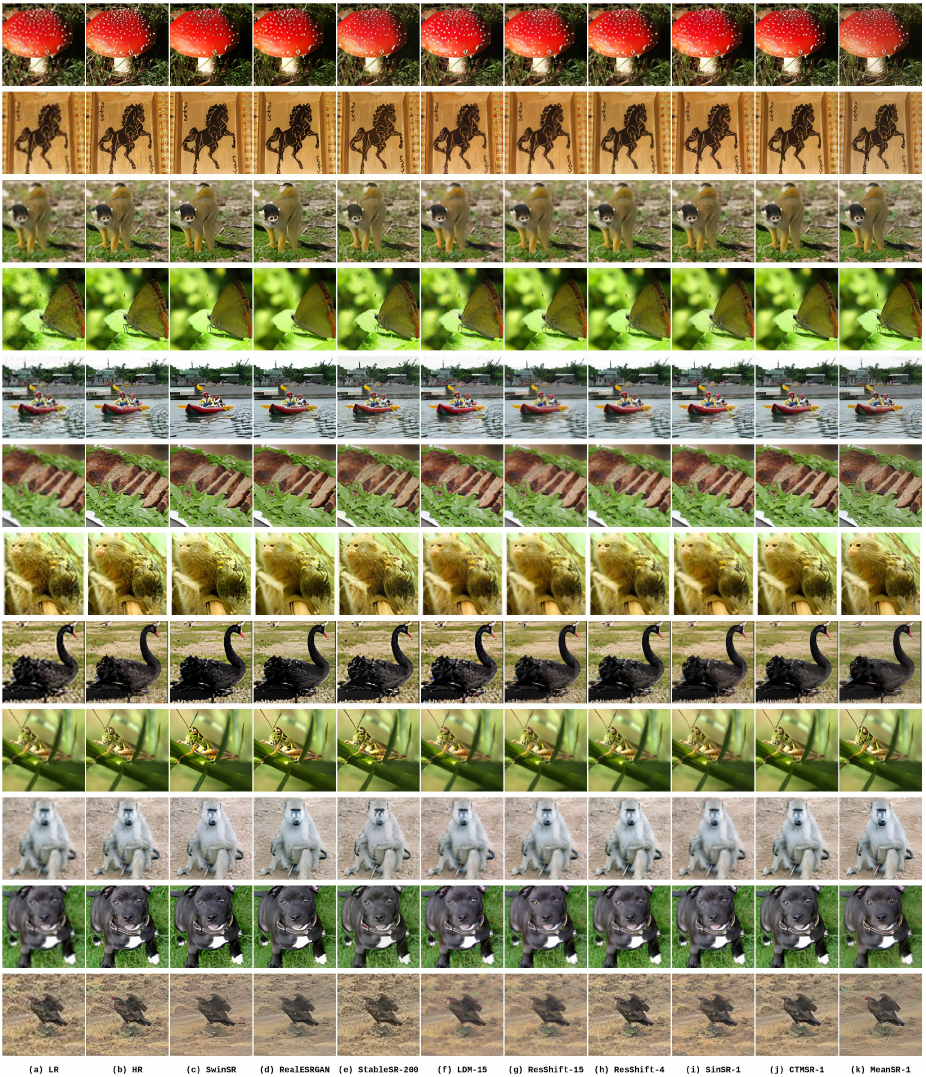}
    \caption{Qualitative comparison on the ImageNet-Test dataset. }
    \label{fig:supp_imagenet}
\end{figure*}

We further provide qualitative comparisons on the synthetic ImageNet-Test benchmark. As shown in Fig.~\ref{fig:supp_imagenet}, MeanSR consistently achieves superior visual quality compared with representative reconstruction-based, diffusion-based, and one-step super-resolution methods.

Compared with reconstruction-oriented methods such as SwinIR and RealESRGAN, MeanSR produces sharper object boundaries and recovers more realistic high-frequency details, alleviating the over-smoothing problem caused by deterministic pixel reconstruction. Compared with multi-step diffusion-based approaches, including StableSR, LDM, and ResShift, MeanSR generates comparable or better perceptual details with significantly fewer sampling steps, demonstrating the effectiveness of explicit restoration trajectory learning.

Furthermore, compared with existing one-step methods such as SinSR and CTMSR, MeanSR produces more natural textures and fewer visual artifacts while maintaining structural fidelity. These results verify that LR-conditioned average-velocity modeling combined with trajectory-level distribution alignment provides effective guidance for one-step perceptual super-resolution.



\end{document}